\documentclass{article}

\PassOptionsToPackage{numbers, compress}{natbib}

\usepackage[preprint]{neurips_2025}

\usepackage[utf8]{inputenc} 
\usepackage[T1]{fontenc}    
\usepackage{hyperref}       
\usepackage{url}            
\usepackage{booktabs}       
\usepackage{amsfonts}       
\usepackage{nicefrac}       
\usepackage{microtype}      
\usepackage{xcolor}         

\usepackage{authblk}  

\usepackage{amsmath} 

\usepackage{multirow, graphicx}
\usepackage{subcaption}

\newcommand{\andrew}[1]{\color{black} #1}

\title{An Explainable Diagnostic Framework for Neurodegenerative Dementias via Reinforcement-Optimized LLM Reasoning}

\author{%
\textbf{Andrew Zamai}$^{1}$\thanks{Correspondence to andrew.zamai@u-bordeaux.fr} \quad \textbf{Nathanaël Fijalkow}$^1$ \quad \textbf{Boris Mansencal}$^1$ \\
\textbf{Laurent Simon}$^1$ \quad \textbf{Eloi Navet}$^1$ \quad \textbf{Pierrick Coupé}$^1$ \\
\vspace{0.5cm}
$^1$Univ. Bordeaux, CNRS, Bordeaux INP, LaBRI, UMR 5800, F-33400 Talence, France\\
}

\begin{document}

\maketitle

\begin{abstract}
The differential diagnosis of neurodegenerative dementias is a challenging clinical task, mainly because of the overlap in symptom presentation and the similarity of patterns observed in structural neuroimaging. To improve diagnostic efficiency and accuracy, deep learning–based methods such as Convolutional Neural Networks and Vision Transformers have been proposed for the automatic classification of brain MRIs. However, despite their strong predictive performance, these models find limited clinical utility due to their opaque decision making.
In this work, we propose a framework that integrates two core components to enhance diagnostic transparency. First, we introduce a modular pipeline for converting 3D T1-weighted brain MRIs into textual radiology reports. Second, we explore the potential of modern Large Language Models (LLMs) to assist clinicians in the differential diagnosis between Frontotemporal dementia subtypes, Alzheimer’s disease, and normal aging based on the generated reports. 
To bridge the gap between predictive accuracy and explainability, we employ reinforcement learning to incentivize diagnostic reasoning in LLMs. Without requiring supervised reasoning traces or distillation from larger models, our approach enables the emergence of structured diagnostic rationales grounded in neuroimaging findings. Unlike post-hoc explainability methods that retrospectively justify model decisions, our framework generates diagnostic rationales as part of the inference process—producing causally grounded explanations that inform and guide the model’s decision-making process. In doing so, our framework matches the diagnostic performance of existing deep learning methods while offering rationales that support its diagnostic conclusions.
\end{abstract}

\section{Introduction}


Neurodegenerative dementias denote a group of disorders characterized by progressive loss of neuronal structure and function, resulting in cognitive, motor, and behavioral impairments~\cite{Lamptey2022}. These conditions typically develop insidiously and worsen over time~\cite{crous-bou2017alzheimer}. Early and accurate diagnosis is therefore critical to slow disease progression and improve patients’ quality of life. However, the differential diagnosis of neurodegenerative dementias—particularly between Alzheimer’s disease (AD), subtypes of Frontotemporal Dementia (FTD), and cognitively normal aging (CN)—remains an open clinical challenge, due to the overlap in symptom presentation and the similarity of patterns observed in structural neuroimaging~\cite{diagnosis_challenges, chouliarasUseNeuroimagingTechniques2023}.
In this work, following~\cite{Harper692}, we focus on structural Magnetic Resonance Imaging (MRI) due to its widespread availability, non-invasive nature, and ability to detect region-specific patterns of cerebral atrophy that are indicative of neurodegeneration.

Existing deep learning approaches, employing either Convolutional Neural Networks (CNNs)~\cite{Hu2021, Ma2020, Nguyen2023} or Vision Transformers (ViTs)~\cite{liDiaMondDementiaDiagnosis2024, nguyen3DTransformerBased2023}, have demonstrated strong performance distinguishing between AD, FTD, and healthy controls from 3D MRI scans. However, despite advances in diagnostic performance, a significant limitation lies in their limited interpretability—the ability to understand the internal mechanics of the model—and their insufficient explainability, that is the capacity to provide human-understandable justifications, or rationales, that clarify why a specific prediction was made. 
Post-hoc explainability methods have been applied to medical imaging~\cite{GRADCAM, nguyenInterpretableDifferentialDiagnosis2022, nguyenDeepGradingMRIbased2023}. 
However, such tools often only indicate \textit{where} the model focused, not \textit{why} it reached a particular diagnosis. Methods like~\cite{nguyenInterpretableDifferentialDiagnosis2022, nguyenDeepGradingMRIbased2023} generate visual heatmaps over MRI scans to indicate which regions influenced the model’s prediction, but they lack semantic attribution—they neither explicitly identify the anatomical regions (e.g., "the hippocampus") nor explain their clinical relevance (e.g., how hippocampal atrophy is indicative of Alzheimer's disease). Furthermore, these methods operate post hoc and play no role in the model’s decision-making process, thereby failing to provide causally grounded justifications.

\begin{figure}[h]
  \centering
  \includegraphics[width=\textwidth]{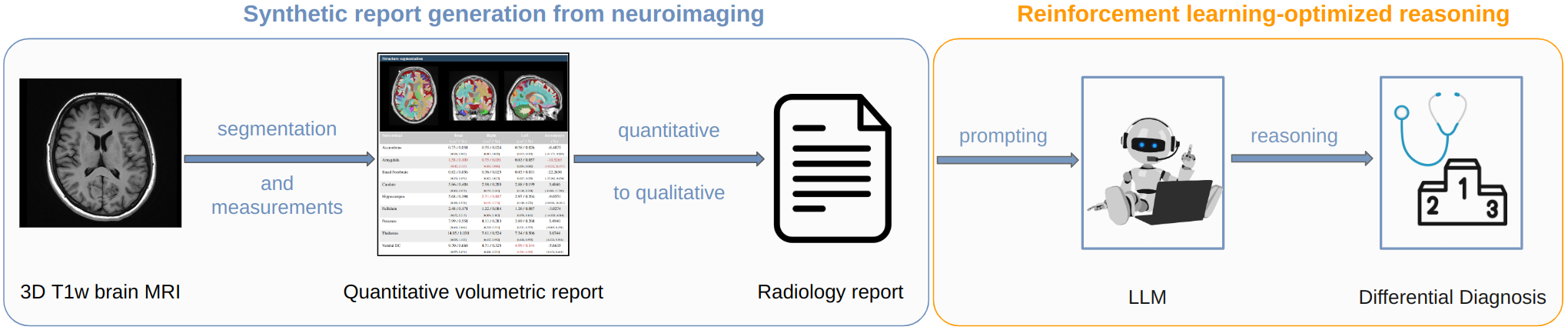}
  \caption{Overview of the proposed framework for the automated differential diagnosis of neurodegenerative dementias. 3D T1-weighted brain MRIs are converted into radiology reports and used to prompt an LLM for detailed diagnostic reasoning and a final ranked list of candidate diagnoses.
  }
  \label{fig:MRI2report_pipeline}
\end{figure}


Our aim in this work is to leverage reasoning-capable Large Language Models (LLMs) to generate rationales that inform and support the diagnostic process, directly addressing key limitations of prior methods.
We present the first comprehensive framework for detailed diagnostic reasoning based on neuroimaging evidence in the differential diagnosis of neurodegenerative dementias\footnote{The framework will be available soon at \url{https://www.volbrain.net/}.}. While vision-to-text models in the medical domain exist~\cite{review_med_vlm}, they typically require massive datasets for training and often fail to capture the fine-grained anatomical details critical for neurological differential diagnosis.
To address this, we introduce a framework that integrates high-resolution segmentation and statistical analysis with the text-based reasoning strengths of LLMs, as illustrated in Figure~\ref{fig:MRI2report_pipeline}.

\paragraph{Synthetic radiology report generation from neuroimaging.}
To enable diagnostic reasoning grounded in neuroimaging evidence, we develop a modular pipeline that transforms T1-weighted 3D MRI scans into textual radiology reports. These reports capture clinically relevant features such as spatial atrophy patterns and anatomical asymmetries, expressed in semantically meaningful terms (e.g., hippocampal atrophy). Unlike end-to-end deep learning solutions~\cite{CT2Rep, HolisticReportGen, hylandMAIRA1SpecialisedLarge2024}, our approach allows each intermediate output to be verified and adjusted for clinical fidelity. We then employ a prompting strategy that guides LLMs to perform differential diagnosis based on these synthetic reports.

\paragraph{Reinforcement learning–optimized reasoning.} To address the lack of training data, we apply Group Relative Policy Optimization (GRPO), a reinforcement learning paradigm recently introduced by DeepSeek~\cite{DeepSeek, DeepSeekMath}, to fine-tune lightweight open-source LLMs, optimizing them to generate diagnostic rationales based on image findings. In the absence of labeled reasoning traces for supervised training, GRPO enables emergent reasoning capabilities without relying on explicit supervision.
Our experiments first evaluate the zero-shot diagnostic capabilities of existing LLMs, providing not only a comparative benchmark of their performance, but also intrinsic validation of our synthetic report generation pipeline. Their zero-shot performance suggests that the generated reports effectively capture clinically relevant information and align with the distributional characteristics of real-world radiology reports likely encountered during pretraining.
Building on this, we apply GRPO-based fine-tuning to lightweight 8B-scale models, enabling them to not only match but often surpass the diagnostic accuracy of much larger models like GPT-4o. Beyond improved diagnostic accuracy, fine-tuned models exhibit detailed and nuanced reasoning grounded in neuroanatomical evidence. Their outputs demonstrate sophisticated behaviors such as hypothesis testing, iterative refinement, and ranked differential diagnoses. Finally, compared to conventional classification-only deep learning solutions, our LLM-based framework achieves competitive diagnostic accuracy while providing transparent, causally grounded rationales that inform and support its diagnostic conclusions.


\section{Related work}
\paragraph{Neuroimaging-based Diagnosis and Post-hoc Explainability Limitations.}
Deep learning models—particularly CNNs~\cite{Hu2021, Ma2020, Nguyen2023} and, more recently, ViTs~\cite{liDiaMondDementiaDiagnosis2024, nguyen3DTransformerBased2023}—have achieved strong performance in structural MRI-based diagnosis. However, these studies primarily focus on distinguishing cognitively normal individuals, AD, and FTD, without explicitly addressing the more challenging task of differentiating between FTD subtypes.
Post-hoc explainability techniques have been utilized in medical imaging~\cite{GRADCAM, nguyenInterpretableDifferentialDiagnosis2022, nguyenDeepGradingMRIbased2023}, but these methods have inherent limitations in this specific context~\cite{GRADCAM_limitations}. These visualizations typically highlight \textit{where} the model focused, without providing insight into \textit{why} a particular diagnosis was made. This is particularly problematic in disorders with overlapping atrophy patterns, where accurate diagnosis depends not just on the presence of atrophy, but on its severity, distribution, and clinical relevance.

\paragraph{LLMs in medicine.}
LLMs have proven effective in encoding medical knowledge~\cite{singhalLargeLanguageModels2023} and supporting various clinical tasks, including medical question answering~\cite{singhalExpertlevelMedicalQuestion2025}, discharging summaries generation~\cite{patelChatGPTFutureDischarge2023, vanveenAdaptedLargeLanguage2024}, electronic health record (EHR) analysis~\cite{jiangHealthSystemscaleLanguage2023}, and text-based differential diagnosis~\cite{mcduffAccurateDifferentialDiagnosis2023}. Domain-adapted models fine-tuned on biomedical corpora—such as PMC-LLaMA, MedAlpaca, BioBERT, and BioGPT~\cite{hanMedAlpacaOpenSourceCollection2025, leeBioBERTPretrainedBiomedical2019, luoBioGPTGenerativePretrained2022, guDomainSpecificLanguageModel2021}—along with multimodal architectures (e.g., Med-Flamingo, LLaVA-Med, Gemini)\cite{moorMedFlamingoMultimodalMedical2023, liLLaVAMedTrainingLarge2023, saabCapabilitiesGeminiModels2024, yangAdvancingMultimodalMedical2024}, are increasingly capable of assisting clinical decision-making tasks. 
LLMs have shown promise in clinical reasoning and explainability. For instance, Savage et al.~\cite{savageDiagnosticReasoningGPT42024} recently demonstrated that GPT-4 can be prompted to produce structured, step-by-step diagnostic reasoning. This approach, along with~\cite{explainableDualInferenceDiffDiagnosis,LLMsClinicalReasoners}, offers physicians a way to assess the plausibility and trustworthiness of LLM-generated predictions.
Recent work by DeepSeek~\cite{DeepSeek} introduced Group Relative Policy Optimization (GRPO), a reinforcement learning method that has powered a new family of LLMs with emergent reasoning capabilities. GRPO shapes the reward signal without relying on labeled preference data or supervised reasoning traces, estimating a group-wise relative advantage across candidate responses, enabling reasoning behaviors to emerge from diverse prompt-response examples alone. This recent work opens the path to the development of reasoning-powered models in medical domain~\cite{medr1}. To our knowledge, no prior work has used GRPO to optimize LLMs for generating diagnostic rationales from radiological reports, especially in the context of differential diagnosis of neurodegenerative dementias.

\section{Approach}
Figure~\ref{fig:MRI2report_pipeline} illustrates the overall architecture of our automated diagnostic framework. The pipeline begins with the processing of a 3D T1-weighted brain MRI using AssemblyNet~\cite{AssemblyNet}, which generates detailed anatomical segmentations organized into a quantitative volumetric report. This data is then translated into a qualitative radiology report, which serves as input for the LLM. In this section, we first describe the process of generating these synthetic radiology reports. We then detail the prompting strategy employed to guide the LLM in interpreting these reports effectively. Finally, we describe the training procedure, with a particular focus on the strategies employed to encourage nuanced and detailed rationale generation through reinforcement-optimized reasoning.

\subsection{From Brain MRI to Text: A Modular Pipeline for Synthetic Report Generation}\label{subsec:brain_to_text}

We propose a modular pipeline capable of converting a 3D T1-weighted brain MRI into a textual radiology report through a series of interpretable intermediate steps. Unlike end-to-end vision-to-text models~\cite{CT2Rep, HolisticReportGen, hylandMAIRA1SpecialisedLarge2024}, our approach preserves the clinical detail essential for the diagnosis of neurodegenerative diseases and offers transparency into each intermediate output.
The pipeline consists of four main stages: (1) fine-grained brain segmentation, (2) volume ratio computation of each anatomical structure, (3) atrophy estimation via normative modeling, and (4) textual report generation. We describe each step in detail below.

\paragraph{(1) MRI segmentation.} 
Fine-grained whole-brain segmentation is obtained using AssemblyNet~\cite{AssemblyNet}, a state-of-the-art deep learning framework designed for high-resolution brain segmentation, employing a multiscale ensemble of 3D U-Net models.
Two assemblies operating at different spatial resolutions enable the model to progressively refine anatomical boundaries and capture detailed structural information. The final output is a detailed segmentation map that includes over 132 brain structures, with specific identification of bilateral elements and detailed left/right segmentation, and a particular focus on cortical, subcortical, and lobar areas—key regions for the diagnosis of neurodegenerative diseases.

\paragraph{(2) Volume ratio computation.} Volume ratios are computed by first measuring the absolute volume of each anatomical region based on the voxel-wise segmentation output; each regional volume is then normalized by the subject's total intracranial volume (ICV) to produce a relative volume ratio. This normalization facilitates the comparison of brain structure sizes between subjects with different brain volumes, accounting for inter-subject variability in brain size.

\begin{figure}[b]
  \centering
  \includegraphics[width=\textwidth]{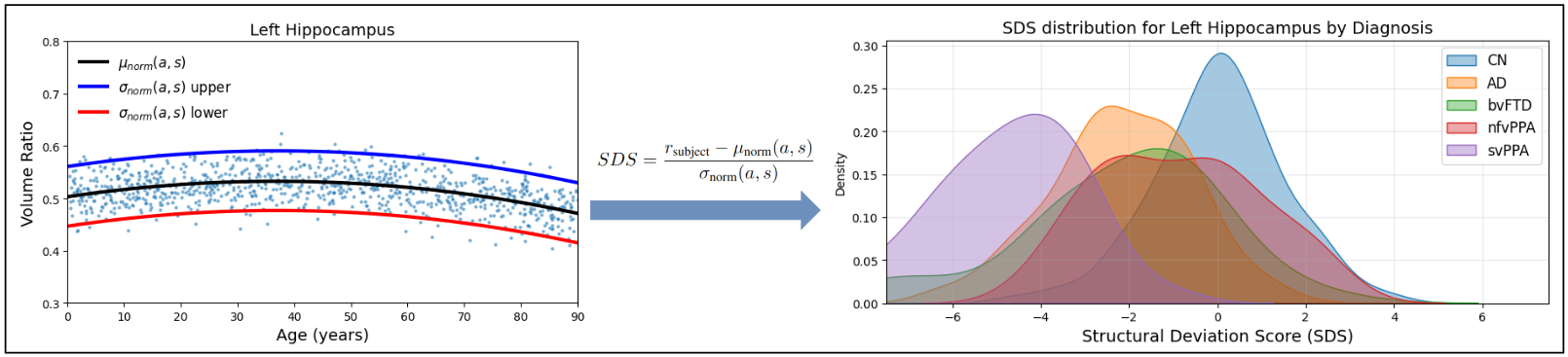}
  \caption{Atrophy estimation via normative modeling. \textbf{Left:} Lifespan curve of left hippocampal volume ratio with normative mean $\mu_{\text{norm}}(a, s)$ (black) and confidence bounds $\pm \sigma_{\text{norm}}(a, s)$ (blue/red). \textbf{Right:} SDS distributions across diagnostic groups, reflecting condition-specific structural deviations.}
  \label{fig:atrophy_estimation}
\end{figure}

\paragraph{(3) Atrophy Estimation via Normative Modeling.} To assess the clinical relevance of volumetric changes in brain structures, our pipeline estimates structural atrophy by comparing an individual’s measured brain volume ratios to normative models that account for age and sex differences. These normative trajectories are derived from the lifespan analysis conducted in \cite{BrainMaturationLifespan}, which provides robust, data-driven volumetric reference curves based on 2,944 high-quality T1-weighted MRI scans from healthy individuals aged 9 months to 94 years.
As depicted in Figure~\ref{fig:atrophy_estimation}, for each brain structure, we compute its Structural Deviation Score (SDS) as $\frac{r_{\text{subject}} - \mu_{\text{norm}}(a, s)}{\sigma_{\text{norm}}(a, s)}$, where $r_{\text{subject}}$ is the subject's measured volume ratio, $\mu_{\text{norm}}(a, s)$ is the expected normative volume ratio for the subject’s age $a$ and sex $s$, and $\sigma_{\text{norm}}(a, s)$ is the corresponding standard deviation from the normative distribution’s 95\% confidence interval. This computation provides a measure of how many standard deviations the subject’s volume ratio deviates from the expected value, where negative SDS indicate smaller-than-expected volumes (atrophy) and positive SDS suggest larger-than-expected volumes (enlargements). As suggested on the right of Figure~\ref{fig:atrophy_estimation}, the magnitude of the score is key for categorizing the severity of these deviations, as it helps distinguish between neurodegenerative diseases with overlapping affected structures.

\begin{figure}[h]
  \centering
  \includegraphics[width=\textwidth]{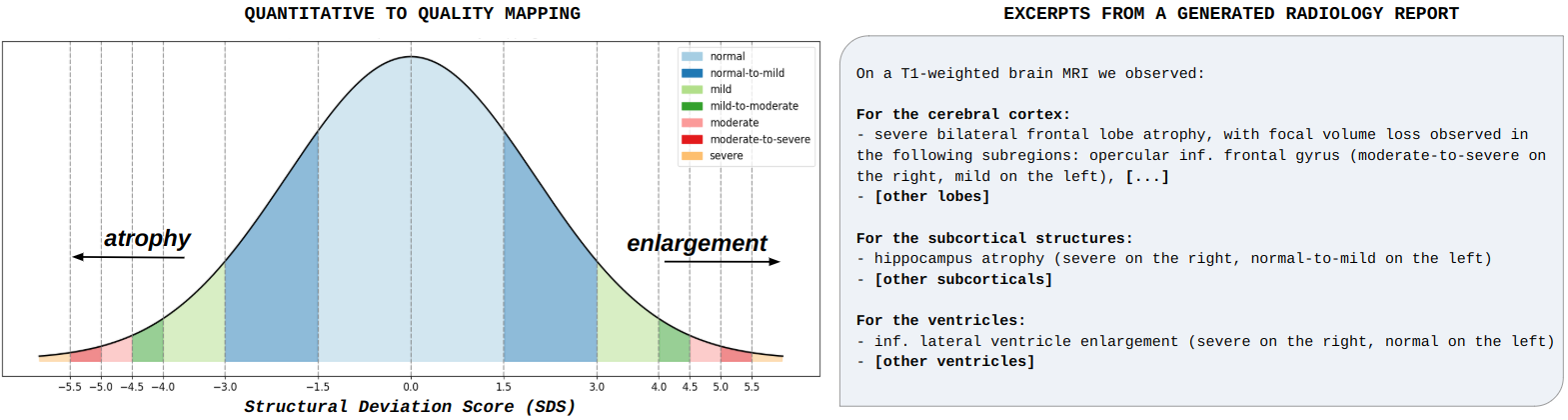}
  \caption{Mapping structural deviation to qualitative severity. \textbf{Left:} Quantitative-to-qualitative conversion of Structural Deviation Scores (SDS) using a seven-point severity scale ranging from severe atrophy to severe enlargement. \textbf{Right:} Example of a (truncated) generated radiology report summarizing anatomical findings by region and hemisphere, using standardized severity descriptors.}
  \label{fig:report_example}
\end{figure}

\paragraph{(4) Radiology Report generation.} The conversion from quantitative volumetric measures to clinically interpretable qualitative descriptions represents a critical component of our pipeline. While the above SDS score provides standardized measurements of deviation from normative reference trajectories, clinicians typically rely on categorical severity assessments—such as mild, moderate, or severe atrophy.
Our pipeline's final stage translates these SDS scores into a descriptive report using a mapping consisting of a seven-point severity scale ranging from \textit{normal} to \textit{severe}, with intermediate gradations (e.g. \textit{normal-to-mild}). As illustrated on the left of Figure~\ref{fig:report_example}, the severity thresholds can be visualized on a Gaussian distribution representing the normative population data, with vertical demarcation lines indicating the boundaries between severity categories. 
The central region of the curve represents volumes within normal limits, while progressively leftward regions correspond to increasing degrees of atrophy severity. Conversely, the right tail of the distribution represents structural enlargement or hypertrophy, which may be particularly relevant for ventricular assessment. This mapping provides an intuitive severity assessments while preserving some of the granularity necessary for differential diagnosis of neurodegenerative conditions with overlapping atrophy patterns. 
Thresholds were chosen based on the statistical meaning of SDS magnitudes, rather than tuning to specific dataset distributions, to prevent overfitting and maintain interpretability. Preliminary zero-shot tests with existing LLMs showed promising results, supporting our decision to retain general-purpose, interpretable thresholds.
The radiology report itself is hierarchically structured, grouping findings by anatomical domain (cortical, subcortical, ventricular). Within each group, regions are sorted by severity, and cortical findings are further differentiated between diffuse lobar atrophy and focal subregional losses—important cues for differential diagnosis. For bilateral structures, we assess both overall and asymmetric volume changes, explicitly noting hemisphere-specific atrophy when present, which is especially relevant for syndromes with lateralized presentations.
By standardizing atrophy descriptions across brain regions using consistent severity terminology, the system enhances clinical communication and supports diagnostic reasoning. In Appendix~\ref{app:full_report}, we provide a full synthetic report alongside a comparison with a human-generated one.

\subsection{Prompting strategy}\label{subsec:prompting_strategy}
\begin{figure}[h!]
  \centering
  \includegraphics[width=\textwidth]{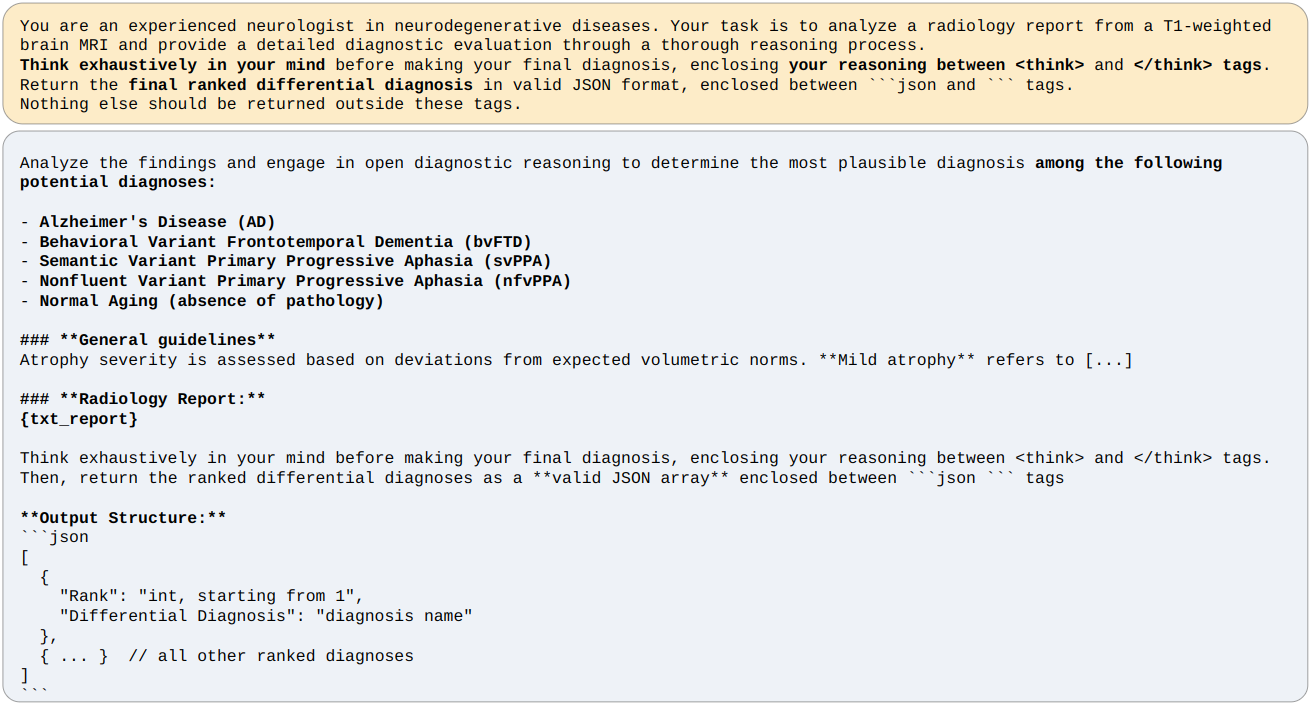}
  \caption{Prompt used to elicit open-ended diagnostic reasoning from MRI reports, ending in a ranked list of differential diagnoses.}
  \label{fig:prompt_template}
\end{figure}

Figure~\ref{fig:prompt_template} illustrates the adopted prompting strategy, designed to elicit an open-ended and thorough diagnostic reasoning based on the neuroimaging findings. The model is instructed to act as a neurologist with expertise in neurodegenerative diseases, tasked with interpreting T1-weighted MRI radiology reports. To encourage deeper engagement with the imaging features, the prompt explicitly requires the model to think exhaustively within \texttt{<think>} tags before committing to a final diagnosis. This intermediate reasoning step encourages detailed examination of regional atrophy patterns, asymmetries, and structural deviations described in the report. 
Finally, the output is structured as a ranked list of differential diagnoses. This format reflects the clinical reasoning process where multiple possibilities are considered and prioritized based on their fit to the observed data. 


{\andrew
To enhance diagnostic stability and consensus at inference, we employ a dual-sampling strategy. First, multiple linguistically varied radiology reports are generated for each brain MRI using sentence templates. Second, the model produces multiple diagnostic predictions for each report via non-deterministic sampling. This approach captures a wider range of interpretations, reducing sensitivity to report phrasing and mitigating LLM inference stochasticity. Final diagnoses are determined by majority vote on the top-ranked differential diagnosis from all aggregated samples, with a supporting reasoning randomly selected from those aligning with the consensus.}

\subsection{Incentivizing diagnostic reasoning with GRPO}
In the absence of labeled reasoning traces for supervised training (SFT), GRPO~\cite{DeepSeekMath, DeepSeek} has been shown to foster emergent reasoning capabilities without relying on explicit supervision or distillation from larger teacher models.
Chosen an LLM as our policy model $\pi_\theta$, where $\theta$ represents its trainable parameters, and a training dataset consisting of tuples $(r, d)$, each comprising a properly converted MRI radiology report $r$ and a gold diagnosis $d$. Each report $r$ is formatted into a prompting query $q$, using the template in Figure~\ref{fig:prompt_template}.
The goal of GRPO is to optimize $\pi_\theta$ such that the generated outputs exhibit human-understandable reasoning that end with a diagnostically accurate prediction. To do so, at each training iteration, for a query $q$ we let the model generate a group of $G$ candidate outputs $\{o_i\}_{i=1}^{G}$ from the current policy $\pi_{\theta_{\text{old}}}$, i.e., the model parameters before the update. Each output $o_i$ contains (1) a diagnostic reasoning trace enclosed between \texttt{<think>}...\texttt{</think>} tags, and (2) a ranked differential diagnosis list in \texttt{JSON} format enclosed in triple backticks \texttt{```json} ... \texttt{```}. For each output $o_i$, GRPO computes a scalar reward $r_i$ using a task-specific reward function (detailed below). It then calculates the \textit{group-relative advantage} $A_i$ for each output as follows:
\[
A_i = \frac{r_i - \mu}{\sigma}, \quad \text{where} \quad \mu = \text{mean}(\{r_1, r_2, \dots, r_G\}), \quad \sigma = \text{std}(\{r_1, r_2, \dots, r_G\}).
\]

Completions with above-average rewards receive amplified policy gradient updates, thereby reinforcing desirable behaviors such as coherent diagnostic reasoning and accurate differential ranking.
The model $\pi_\theta$ is updated by maximizing $\mathcal{J}_{\text{GRPO}}$, as further detailed in the original work by DeepSeek~\cite{DeepSeek}.


We design our task-specific reward function to consist of two terms, a format and accuracy reward.
\paragraph{Format reward.} Each completion is scored on a 1.0-point scale, composed of four equally weighted components (0.25 points each): (1) the presence of a <think>...</think> reasoning block followed immediately by a JSON output ensures proper structural delimitation; (2) a single well-formed JSON block enclosed in triple backticks guarantees parseability; (3) the ability to extract the top-ranked diagnosis from the JSON confirms correct output formatting; and (4) the inclusion of all $K$ expected diagnostic categories, matched via regular expressions, checks for class coverage and no other diseases considered. If multiple diagnoses are assigned the top rank, the total reward is capped at 0.25 to penalize ambiguity.
    
\paragraph{Accuracy reward.} We assign a binary reward by comparing the model’s top-ranked diagnosis to the ground truth. The predicted diagnosis is extracted from the JSON output, mapped to a class ID using a regex-based scheme, and compared against the gold label. A reward of 1.0 is given for a correct match, and 0.0 otherwise, incentivizing clinically accurate predictions.



\section{Experiments}
Our experiments aim to assess: (1) the zero-shot diagnostic performance of existing LLMs in interpreting our synthetic reports—and, conversely, how well these reports align with their pretrained knowledge and typical report distributions; (2) the impact of GRPO fine-tuning on the emergence of diagnostic reasoning; and (3) how our framework compares to established deep learning classifiers trained directly on brain MRI data.

\subsection{Experimental setup}

\paragraph{Datasets.}
Table~\ref{tab:diagnosis_distribution} summarizes the diagnostic distribution across training, validation, and test sets for all 615 participants included in this study. The data were aggregated from two major sources: the Alzheimer’s Disease Neuroimaging Initiative (ADNI)~\cite{ADNI} and the Neuroimaging Initiative for Frontotemporal Lobar Degeneration (NIFD)\footnote{Available at \url{https://ida.loni.usc.edu/}}. Cases from NIFD included behavioral variant frontotemporal dementia (bvFTD), non-fluent variant primary progressive aphasia (nfvPPA), and semantic variant primary progressive aphasia (svPPA). Alzheimer’s disease (AD) cases were drawn from ADNI, and cognitively normal (CN) controls from both datasets. 
{\andrew Differentiating between FTD subtypes using only structural neuroimaging is challenging, due to the heterogeneity in atrophy patterns within each subtype, the potential for overlapping regions of neurodegeneration across subtypes (e.g., anterior temporal involvement in both svPPA and some bvFTD, or fronto-insular atrophy in both nfvPPA and some bvFTD), particularly in early stages, and the tendency for atrophy to become more widespread with disease progression, further obscuring initial distinctions. Limited samples—particularly for nfvPPA and svPPA—exacerbate these challenges for robust model training and evaluation. To mitigate this, we applied stratified splitting to preserve class distributions and deliberately allocated more data to the test set—at the expense of training set size—in order to retain sufficient samples for reliable evaluation.}

\begin{table}[ht]
\centering
\caption{Diagnostic distribution across splits for all 615 participants included in this study.}
\centering
\begin{tabular}{r|ccccc|c}
\toprule
\textbf{Split} & \textbf{CN} & \textbf{AD} & \textbf{bvFTD} & \textbf{nfvPPA} & \textbf{svPPA} & \textbf{Total} \\
\midrule
Train       & 160 & 75 & 38 & 20 & 17 & 310 \\
Validation  & 62  & 30  & 14 & 8 & 8 & 122 \\
Test        & 94  & 44  & 22 & 12 & 11 & 183 \\
\bottomrule
\end{tabular}
\label{tab:diagnosis_distribution}
\end{table}

\paragraph{Models and training details.}\label{subsec:exp_setup}
We evaluate a diverse set of LLMs encompassing a range of model families, parameter scales, and reasoning capabilities. 
Specifically, we consider the GPT-4o model~\cite{openaiGPT4TechnicalReport2024}, renowned for its strong general-purpose performance, as well ranked highly on the Open Medical LLM Leaderboard~\footnote{\url{https://huggingface.co/spaces/openlifescienceai/open_medical_llm_leaderboard}}. From the top of this leaderboard, we also include Llama3-OpenBioLLM-70B~\cite{OpenBioLLMs}, a domain-specialized model based on the LLaMA-3 70B architecture and fine-tuned for medical tasks.
In addition to these, we assess several open-source generalist models across different model families. From the LLaMA family, we include Llama-3.3-Instruct models at 70B and 8B parameter scales~\cite{llama3herdmodels}. From the Qwen family, we evaluate Qwen-2.5-Instruct-7B~\cite{qwen25technicalreport} and the recent natively reasoner Qwen-3-8B~\cite{yang2025qwen3technicalreport}.
Finally, we evaluate reasoning-augmented models trained via GRPO as introduced by DeepSeek~\cite{DeepSeek}. Specifically, we consider two distilled variants of these reasoner models aligned to the LLaMA architecture, with parameter sizes of 70B and 8B.

Resource constraints limited our fine-tuning efforts to the smaller 8B variants. Notably, GRPO training is computationally intensive; our setup utilized four NVIDIA H100 80GB GPUs, allocating three for training and one for completions generation using the vLLM framework~\cite{vllm}.
To enable efficient fine-tuning, we employed Low-Rank Adaptation (LoRA)~\cite{lora}, using a rank $r = 16$ and scaling factor $\alpha = 32$, targeting the query, key, and value projection weights in the attention layers.
Specific to GRPO training, we followed established configurations from~\cite{liu2025understandingr1zeroliketrainingcritical} and~\cite{DeepSeek}. We set the maximum generation length to 3000 tokens and generated $G = 6$ completions per query, with a sampling temperature of 0.9 to encourage exploration. In accordance with~\cite{liu2025understandingr1zeroliketrainingcritical}, we adopted a max-completion-length averaging scheme for the loss computation, mitigating the bias on completions length.
However, contrary to their recommendation, we retained deviation scaling in the advantage estimation, as removing it consistently degraded training performance in our setting. To ensure stable updates and accommodate a diverse query set across multiple classes, we used a gradient accumulation step size of 64 and a conservative learning rate of $5 \times 10^{-5}$, preventing abrupt shifts in model behavior.
Finally, we preserved the default GRPO $\epsilon$ parameter at 0.2, as in~\cite{DeepSeek}, but found that lowering $\beta$ to 0.005 was crucial. Setting $\beta = 0$ caused the model to produce incoherent outputs and deviate from its pre-trained domain knowledge, whereas larger values excessively suppressed the advantage estimation term, limiting effective learning.

\subsection{Results}
\paragraph{Off-the-shelf LLMs.}
Table~\ref{tab:results} presents the zero-shot diagnostic performance of the models selected in Subsection~\ref{subsec:exp_setup}. 
Using the inference prompting strategy outlined in Subsection~\ref{subsec:prompting_strategy}, we generate $3$ synthetic reports per brain MRI, each followed by $3$ independent diagnostic predictions, yielding a total of $9$ candidate outputs per case. The final diagnosis is obtained via majority voting, promoting a more stable and consensus-driven decision. In terms of M-F1, GPT-4o demonstrates the strongest overall zero-shot performance, confirming to be a top-tier generalist model. Among the open-source 70B models, the DeepSeek-R1-Distill-Llama achieves the best results, highlighting the effectiveness of GRPO-style reasoning also for clinical tasks. Notably, Llama3-OpenBioLLM, despite being domain-specialized, performs worse than its base model, likely due to fine-tuning on radiology reports predominantly associated with pathological cases, introducing a bias toward dementia prediction.
Among the 8B models, DeepSeek-R1-Distill-Llama and LLaMA-3.1-Instruct demonstrate strong zero-shot performance, even surpassing some 70B models, while their 3B variants, lacking sufficient pre-trained domain knowledge at that scale, were excluded for poor performance.
These results not only provide a comparative benchmark of the diagnostic capabilities of each model, but also offer an intrinsic validation of the proposed synthetic report generation pipeline, indicating that the generated reports effectively capture clinically relevant information and exhibit distributional characteristics consistent with real-world radiology reports likely encountered during pre-training.

\begin{table*}[h]
    \centering
    \caption{Diagnostic performance of off-the-shelf LLMs and GRPO fine-tuned 8B variants. 
    Models marked with~\textdagger~are reasoning models or distilled from them.}
    \small
    \resizebox{\textwidth}{!}{
    \begin{tabular}{r|c|ccccc|cc}
        \toprule
        \multirow{2}{*}{\textbf{Model}} & \multirow{2}{*}{\textbf{Params}} & \multicolumn{5}{c|}{\textbf{\textit{class-wise} F1}} & \multirow{2}{*}{\textbf{BACC}} & \multirow{2}{*}{\textbf{$\textbf{M}$-F1}} \\
        \cmidrule(lr){3-7}
        & & CN & AD & bvFTD & nfvPPA & svPPA & & \\
        \toprule
        \multicolumn{9}{c}{\hspace{3.5cm}\textit{Zero-shot}} \\
        \midrule
        \textbf{gpt-4o} & - & 70.81 & 51.76 & 51.52 & 19.05 & 48.48 & 55.46 & 48.32 \\
        \midrule
        \textbf{Llama-3.3-Instruct} & 70B & 59.57 & 43.37 & 44.71 & 0.00 & 42.11 & 48.94 & 37.95 \\
        \textbf{Llama3-OpenBioLLM} & 70B & 43.55 & 44.44 & 44.74 & 0.00 & 36.36 & 49.38 & 33.82 \\
        \textbf{DeepSeek-R1-Distill-Llama~\textdagger} & 70B & 64.38 & \textbf{59.93} & 48.19 & 0.00 & 31.25 & 48.18 & 39.55 \\
        \midrule
        \textbf{DeepSeek-R1-Distill-Llama~\textdagger} & 8B & 72.05 & 53.19 & 38.46 & 11.76 & 25.00 & 42.64 & 40.09 \\
        
        \textbf{LlaMA-3.1-Instruct} & 8B & 64.94 & 31.75 & 54.05 & 15.38 & 36.73 & 53.06 & 40.57 \\
        \textbf{Qwen-2.5-Instruct} & 7B & 47.24 & 47.41 & 36.84 & 0.00 & 0.00 & 33.66 & 26.23 \\
        \textbf{Qwen-3~\textdagger} & 8B & 60.56 & 54.95 & 39.13 & 19.05 & 10.00 & 42.03 & 36.74 \\
        
        \midrule
        \multicolumn{9}{c}{\hspace{3.5cm}\textit{Our GRPO fine-tuned models}} \\
        \midrule

        \textbf{DeepSeek-R1-Distill-Llama GRPO~\textdagger} & 8B & 84.16 & 51.43 & 70.59 & 11.76 & \textbf{80.00} & 62.48 & 59.55 \\
        \textbf{LlaMA-3.1-Instruct GRPO~\textdagger} & 8B & \textbf{85.86} & 53.33 & \textbf{73.17} & 41.67 & 71.43 & 67.33 & \textbf{65.09} \\
        \textbf{Qwen-2.5-Instruct GRPO~\textdagger} & 7B & 78.36 & 49.44 & 58.06 & 0.00 & 42.42 & 52.89 & 45.66 \\
        \textbf{Qwen-3 GRPO~\textdagger} & 8B & 83.33 & 46.88 & 66.67 & \textbf{48.00} & 64.52 & \textbf{68.38} & 61.88 \\

        \bottomrule
    \end{tabular}
    }
    
    \label{tab:results}
\end{table*}

\begin{figure}[h!]
  \centering
  \includegraphics[width=\textwidth]{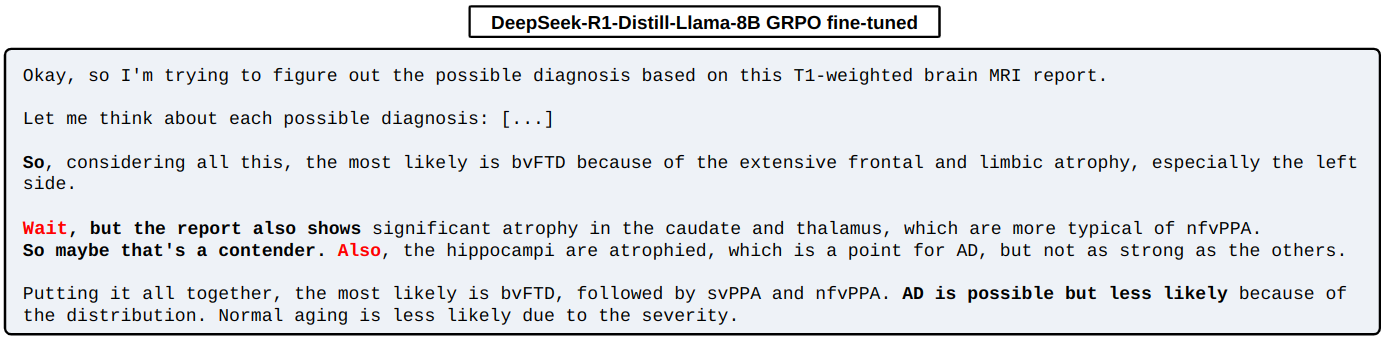}
  \caption{Excerpts from the DeepSeek-R1-Distill-Llama-8B-GRPO model. The responses exhibit properties such as evidence-based hypothesis testing, non-linear reasoning, and detailed understanding of expected anatomical regions and atrophy severity.}
  \label{fig:output_extracts}
\end{figure}

\paragraph{GRPO Fine-Tuned LLMs.} We extend Table~\ref{tab:results} by reporting the diagnostic performance of the 8B models fine-tuned via GRPO, as detailed in Subsection~\ref{subsec:exp_setup}. Remarkably, without any supervised reasoning traces or distillation from larger models, GRPO enables the emergence of detailed, evidence-based diagnostic reasoning that contributes to improved diagnostic accuracy.
Due to the length of the generated outputs, in Figure~\ref{fig:output_extracts} we present only excerpts from the DeepSeek-R1-Distill-Llama-8B model. Qualitative analysis of these outputs reveals several key reasoning behaviors. First, the models engage in explicit ``hypothesis testing'', systematically evaluating each candidate diagnosis by weighing supporting and opposing imaging features. This promotes balanced consideration across differential diagnoses rather than premature commitment. Second, the models demonstrate ``non-linear reasoning'', often revisiting and refining earlier conclusions as additional evidence is considered. Third, responses typically conclude with a ranked list of differential diagnoses, reflecting varying degrees of confidence rather than a single-label decision. 
The rationales exhibit a high degree of anatomical specificity, referencing expected neuroanatomical atrophies and capturing distribution patterns that reflect known disease profiles, including severity and asymmetry. Finally, we observe that output length and detail correlate with case complexity. For straightforward cases (e.g., cognitively normal scans or reports with hallmark disease features), the model produces concise justifications. In contrast, challenging cases elicit significantly longer and more elaborate reasoning—sometimes up to three times longer.
Further insights into training dynamics and full diagnostic reasoning examples are provided in Appendix~\ref{app:grpo_training} and Appendix~\ref{app:full_responses}, respectively.


\paragraph{Comparison with classification-only approaches.} We compare our LLM-based diagnostic framework with existing deep-learning classification methods training directly on brain MRIs. Nguyen \textit{et al.}~\cite{nguyen3DTransformerBased2023} demonstrated that 3D transformer-based models achieve strong predictive performance, outperforming previous approaches~\cite{maDifferentialDiagnosisFrontotemporal2020a, Hu2021}. While these prior works focus on distinguishing between AD, FTD, and CN cases, we extended Nguyen \textit{et al.}’s framework to include FTD subtypes.
In line with their setup, we also evaluate a Support Vector Machine (SVM) classifier trained on the Structural Deviation Scores computed as detailed in Subsection~\ref{subsec:brain_to_text}. Further implementation and training details are provided in Appendix~\ref{app:dl_training_details}.
Table~\ref{tab:DL_comparion} presents the diagnostic performance of the ViT and SVM classifiers, compared with our best-performing zero-shot LLM and best GRPO fine-tuned variants. 
{\andrew While GradCAM-based post hoc visualizations in~\cite{nguyen3DTransformerBased2023} offer some interpretability by highlighting image regions that influence model predictions, they lack semantic attribution and clinical contextualization. In contrast, our LLM-based framework achieves comparable diagnostic performance while producing transparent, human-readable rationales that explicitly reference neuroanatomical structures and articulate their relevance to the differential diagnosis.}

\begin{table*}[h]
    \centering
    \caption{Diagnostic performance comparison between our LLM-based framework and existing classification-only deep learning approaches.}
    \small
    \resizebox{\textwidth}{!}{
    \begin{tabular}{r|c|ccccc|cc}
        \toprule
        \multirow{2}{*}{\textbf{Model}} & \multirow{2}{*}{\textbf{Params}} & \multicolumn{5}{c|}{\textbf{\textit{class-wise} F1}} & \multirow{2}{*}{\textbf{BACC}} & \multirow{2}{*}{\textbf{$\textbf{M}$-F1}} \\
        \cmidrule(lr){3-7}
        & & CN & AD & bvFTD & nfvPPA & svPPA & & \\
        \toprule
        \multicolumn{9}{c}{\hspace{2.5cm}\textit{Classification only}} \\
        \midrule
        \textbf{3D-Vision Transformer} & 64M & \textbf{88.44} & \textbf{72.50} & 62.50 & 12.50 & 78.26 & 63.57 & 62.84 \\
        \textbf{SVM atrophies} & - & 86.43 & 69.66 & \textbf{73.91} & 0.00 & \textbf{84.21} & 62.39 & 62.85 \\
            
        \midrule
        \multicolumn{9}{c}{\hspace{2.5cm}\textit{LLMs providing diagnostic reasoning}} \\
        \midrule
        
        \textbf{gpt-4o} & - & 70.81 & 51.76 & 51.52 & 19.05 & 48.48 & 55.46 & 48.32 \\
        \textbf{DeepSeek-R1-Distill-Llama GRPO~\textdagger} & 8B & 84.16 & 51.43 & 70.59 & 11.76 & 80.00 & 62.48 & 59.55 \\
        \textbf{LLaMA-3.1-Instruct GRPO~\textdagger} & 8B & 85.86 & 53.33 & 73.17 & 41.67 & 71.43 & 67.33 & \textbf{65.09} \\
        \textbf{Qwen-3 GRPO~\textdagger} & 8B & 83.33 & 46.88 & 66.67 & \textbf{48.00} & 64.52 & \textbf{68.38} & 61.88 \\
        
        \bottomrule
    \end{tabular}
    }
    
    \label{tab:DL_comparion}
\end{table*}

\section{Conclusions}
We introduced a modular framework for the differential diagnosis of neurodegenerative dementias that combines high-resolution MRI analysis, synthetic radiology reporting, and LLM-based reasoning. 
By shifting from post hoc explanations to inference-time diagnostic reasoning, our method provides anatomically grounded rationales, offering physicians a way to access the plausibility and trustworthiness of LLM-generated predictions.
Fine-tuning lightweight LLMs via reinforcement learning with Group Relative Policy Optimization (GRPO), we demonstrate that coherent diagnostic reasoning can be achieved without requiring supervised reasoning traces.

This work demonstrates the promise of using reasoning models in clinical contexts—particularly for complex, multi-hypothesis tasks such as differential diagnosis. We take an important first step in this direction, opening avenues for future systems that combine data-driven prediction with structured and transparent reasoning.

\clearpage
\section*{Acknowledgments}
This work benefited from the support of the project HoliBrain funded by the French National Research Agency (ANR-23-CE45-0020-01) and  the prematuration project ChatvolBrain funded by the CNRS. Moreover, this project is supported by the Precision and global vascular brain health institute funded by the France 2030 investment plan as part of the IHU3 initiative (ANR-23-IAHU- 0001). Finally, this study received financial support from the French government in the framework of the University of Bordeaux's France 2030 program / RRI "IMPACT and the PEPR StratifyAging. This work was also granted access to the HPC resources of IDRIS under the allocation 2022-AD011013848R1 made by GENCI. 

\bibliographystyle{unsrtnat}
\bibliography{references}


\appendix
\section{Example of a Complete Synthetic Report and Human Comparison}\label{app:full_report}
We provide an illustrative example of a complete synthetic radiology report generated from a 3D T1-weighted MRI scan using the pipeline detailed in Subsection~\ref{subsec:brain_to_text}. The input scan was obtained from an open-access case on Radiopaedia.org~\footnote{\url{https://radiopaedia.org/cases/frontotemporal-dementia-behavioural-variant-2}}, enabling a direct comparison with expert-curated findings authored by neuroradiologist Dr. Frank Gaillard (founder of Radiopaedia).

\begin{figure}[h]
\centering
\includegraphics[width=\textwidth]{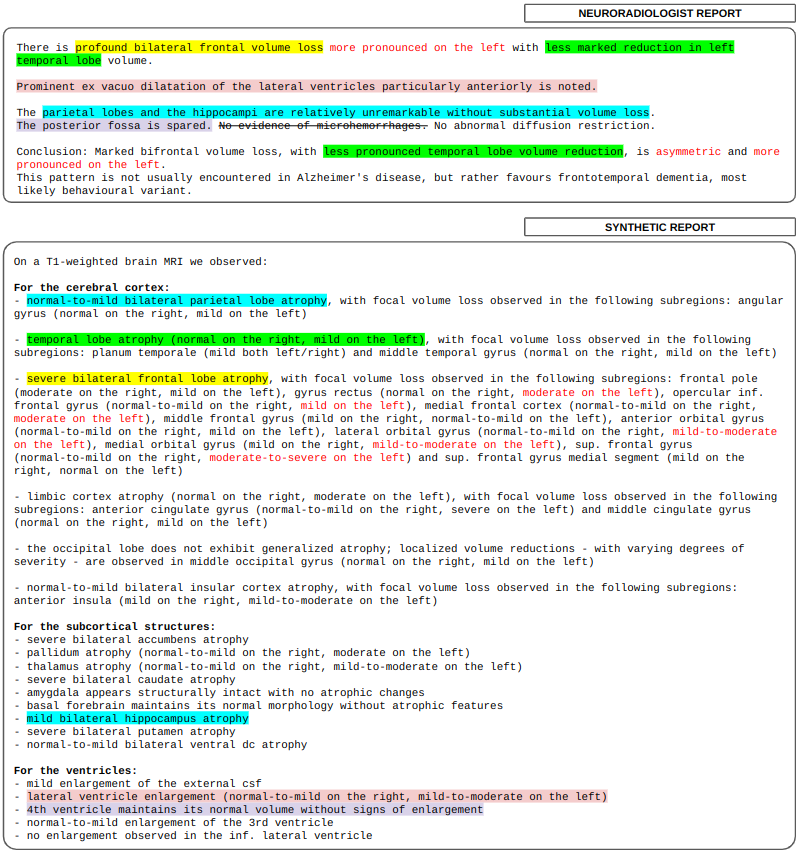}
\caption{Comparison of expert-written (top) and synthetic (bottom) radiology reports for the same T1-weighted MRI scan. Colored highlights indicate overlapping descriptions of atrophic patterns.}
\label{fig:human_syn_report}
\end{figure}

Figure~\ref{fig:human_syn_report} displays the expert-written report (top) and the corresponding synthetic report generated by our model (bottom). To aid visual inspection, we highlight areas of overlap in atrophy pattern descriptions using color coding.

We observe that while both reports capture key neurodegenerative features consistent with the clinical picture—in this case, a diagnosis of behavioral variant frontotemporal dementia (bvFTD)—their reporting styles differ. The synthetic report is more exhaustive and systematically structured: it enumerates a wide range of anatomical regions, including both affected and unaffected areas. In contrast, the expert human report adopts a more concise, diagnosis-driven narrative, focusing selectively on findings most relevant to the suspected pathology. This reflects typical clinical reporting practice, where radiologists tailor descriptions to guide differential diagnosis rather than provide exhaustive anatomical reviews. Nonetheless, despite stylistic differences, the synthetic report successfully captures all key anatomical patterns relevant to the diagnosis, demonstrating its effectiveness in supporting the subsequent differential diagnosis task.

\clearpage
\section{Insight into GRPO Training Dynamics}
\label{app:grpo_training}
\begin{figure}[h]
\centering
\begin{subfigure}[b]{0.48\textwidth}
\centering
\includegraphics[width=\textwidth]{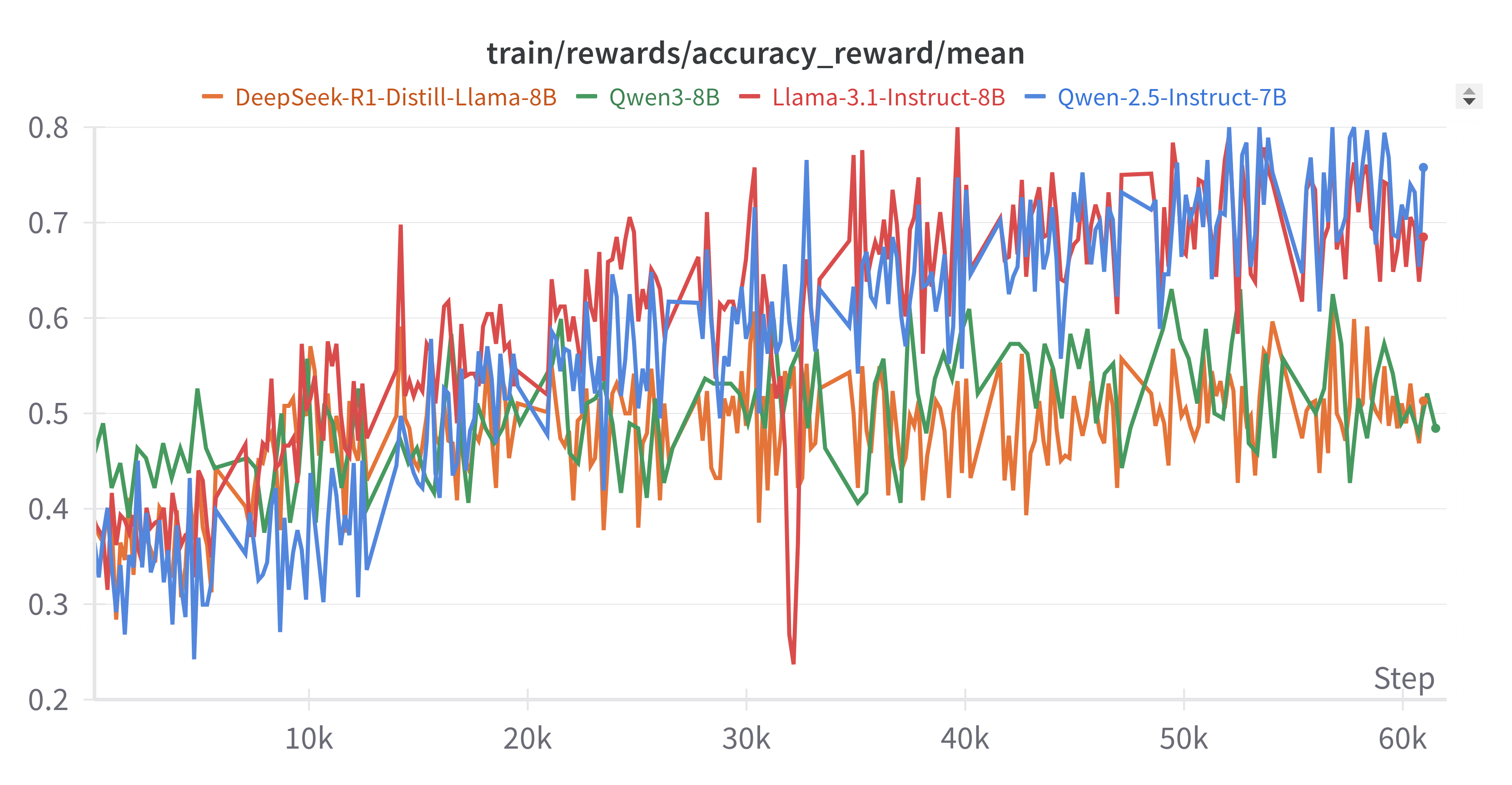}
\caption{Training set: Mean diagnostic accuracy reward}
\end{subfigure}
\hfill
\begin{subfigure}[b]{0.48\textwidth}
\centering
\includegraphics[width=\textwidth]{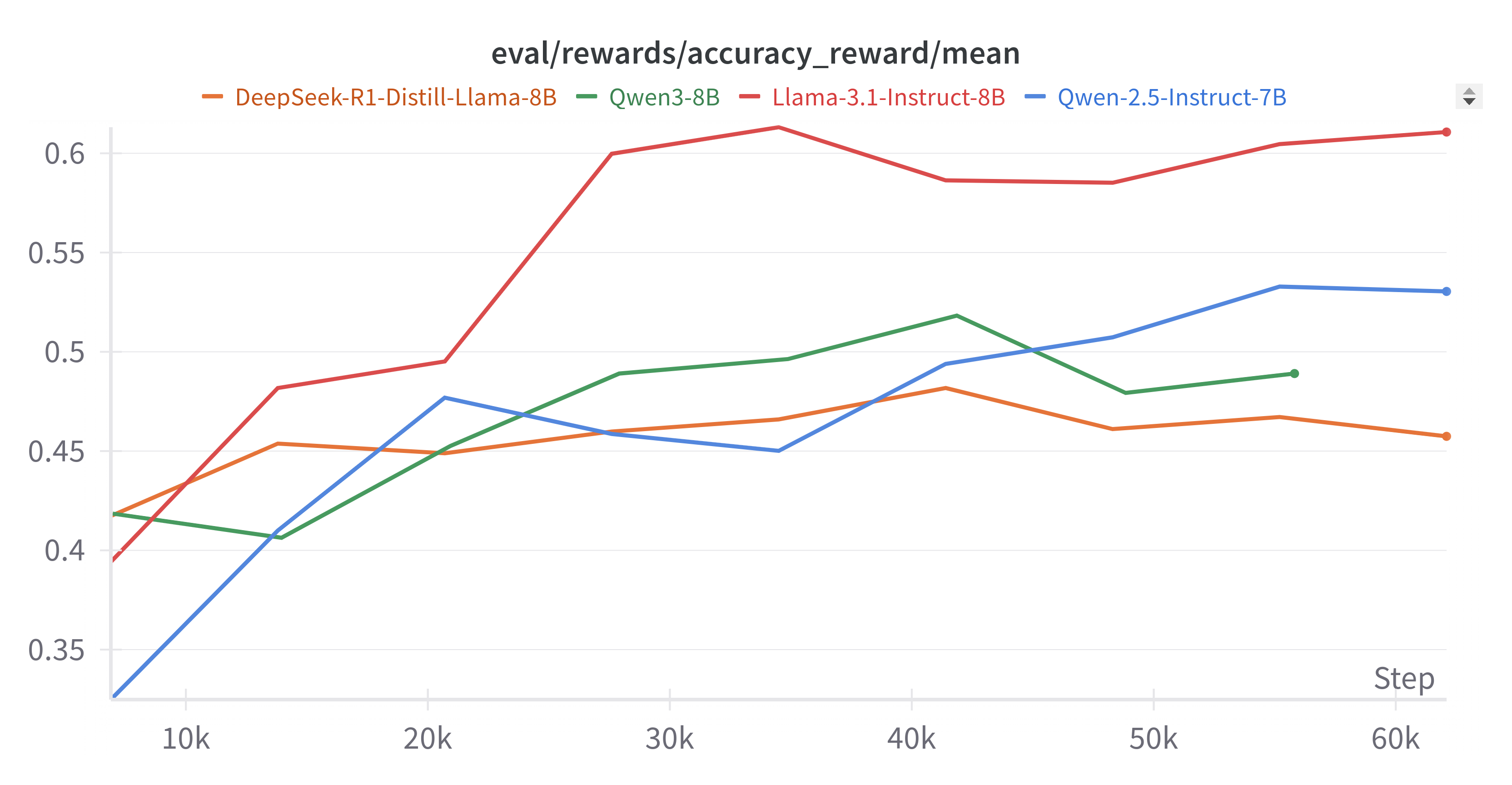}
\caption{Validation set: Mean diagnostic accuracy reward}
\end{subfigure}
\caption{Progression of diagnostic accuracy during the initial 60K steps of GRPO training across training and validation sets.}
\label{fig:GRPO_training}
\end{figure}

\begin{figure}[h]
\centering
\begin{subfigure}[b]{0.48\textwidth}
\centering
\includegraphics[width=\textwidth]{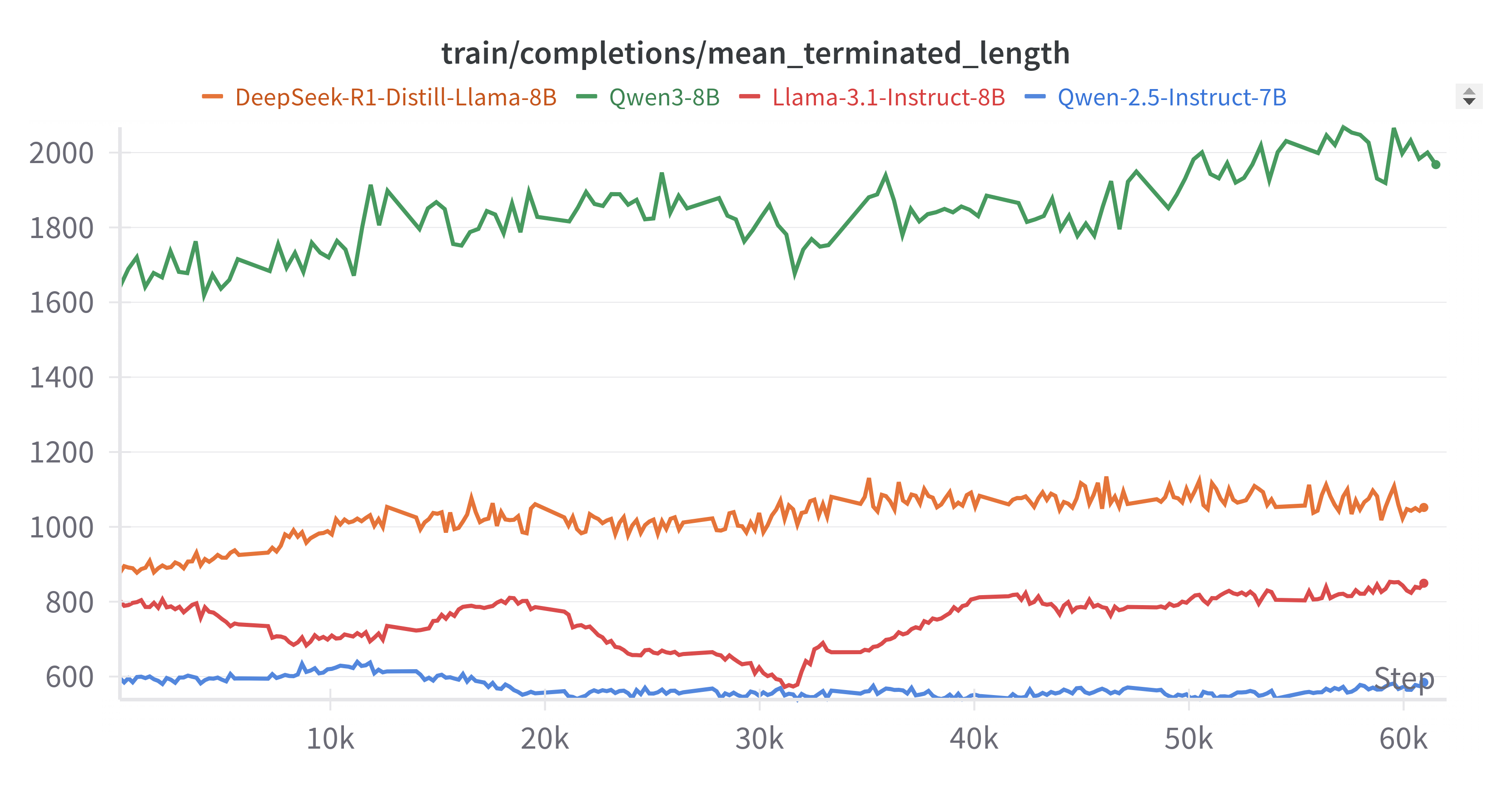}
\caption{Mean generated response length (tokens)}
\end{subfigure}
\hfill
\begin{subfigure}[b]{0.48\textwidth}
\centering
\includegraphics[width=\textwidth]{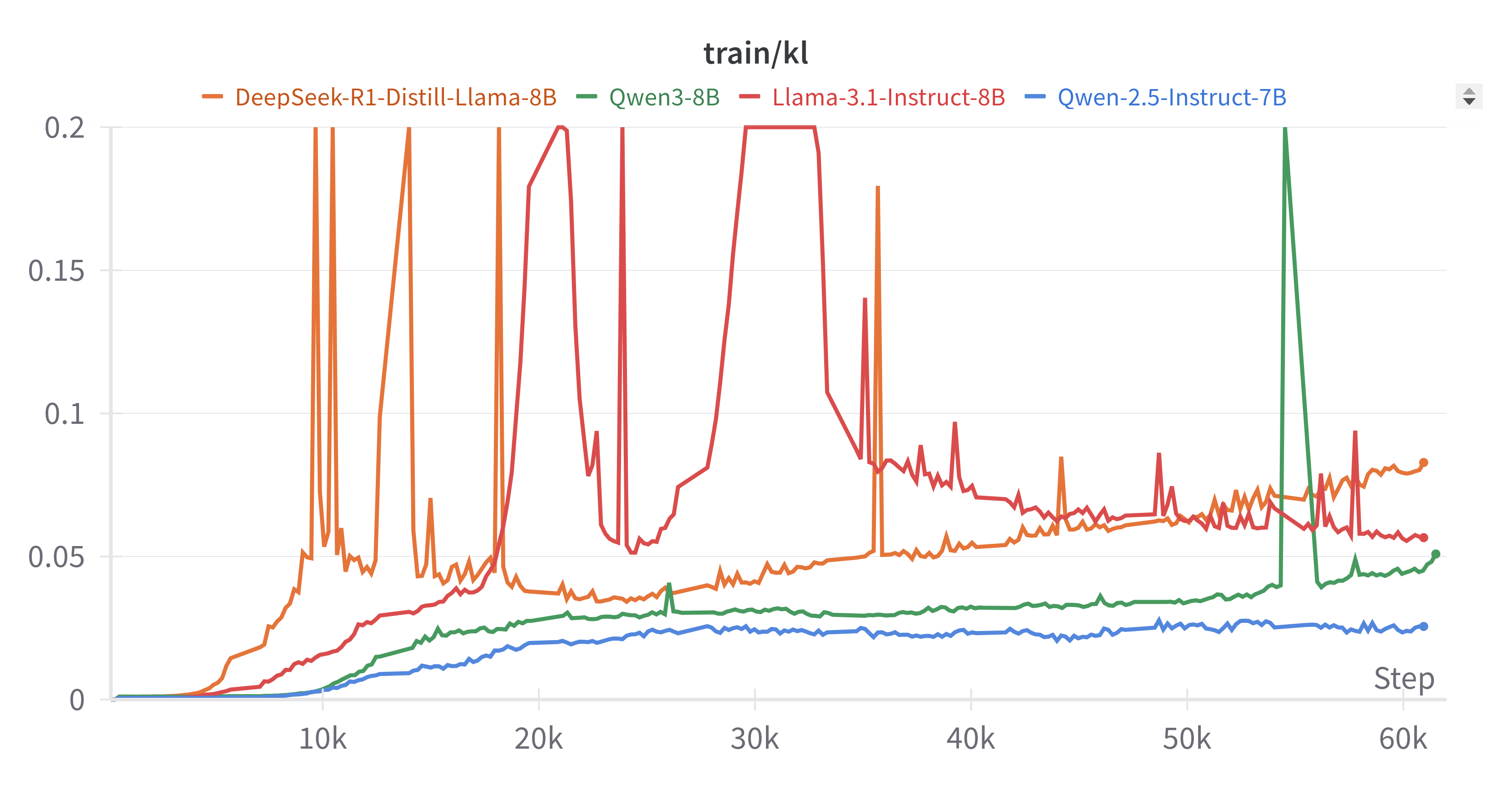}
\caption{KL divergence relative to the original model}
\end{subfigure}
\caption{Evolution of mean response length and KL divergence over the first 60K steps of GRPO training.}
\label{fig:GRPO_training_2}
\end{figure}

This appendix provides additional insight into the training dynamics of the GRPO-fine-tuned 8B models. In Figure~\ref{fig:GRPO_training}, we visualize key performance metrics across training iterations. The left panel illustrates the progression of the mean diagnostic accuracy reward on the training set, while the right panel shows the corresponding performance on the validation set. 
All models exhibit a consistent increase in diagnostic accuracy as training progresses, with a similar upward trend on the validation set, highlighting the effectiveness and generalizability of the GRPO optimization process.
Figure~\ref{fig:GRPO_training_2} presents the evolution of two additional metrics. The left panel depicts the mean length of generated completions (i.e., the number of generated tokens). Interestingly, the trend in response length varies across models. For instance, the Qwen3-8B and DeepSeek-Llama-8B models show a clear upward trend in maximum response length (more marked in the first), aligning with observations in ~\cite{yang2025qwen3technicalreport, DeepSeek}. This suggests a growing capacity for analysis, reasoning refinement, and even self-doubt, as the model develops more nuanced diagnostic justifications.
In contrast, models that are not natively trained for reasoning, such as Qwen-2.5-Instruct and LLaMA-3.1-Instruct, maintain a relatively stable response length throughout training. This stability could suggest early convergence in reasoning style, with responses remaining short and focused—reflecting a compressed but effective form of justification.
Finally, the right panel of Figure~\ref{fig:GRPO_training_2} shows the evolution of the KL divergence term relative to the original model. We observe that all models diverge from their initial response distributions over the course of training. This indicates that GRPO effectively reshapes the model's output behavior while maintaining human-readability through this regularization KL penalty.

\clearpage
\section{Full Diagnostic Reasoning Examples}
\label{app:full_responses}
We present complete diagnostic outputs generated by GRPO-trained models on a randomly sampled case of behavioral variant frontotemporal dementia (bvFTD). These examples highlight distinct reasoning styles that emerged during training. The LLaMA and Qwen models produce concise, focused justifications, emphasizing key diagnostic features with minimal elaboration. In contrast, the DeepSeek and Qwen-3 model exhibits a more expansive diagnostic narrative, engaging in lengthier analysis.

\begin{figure}[h]
\centering
\includegraphics[width=\textwidth]{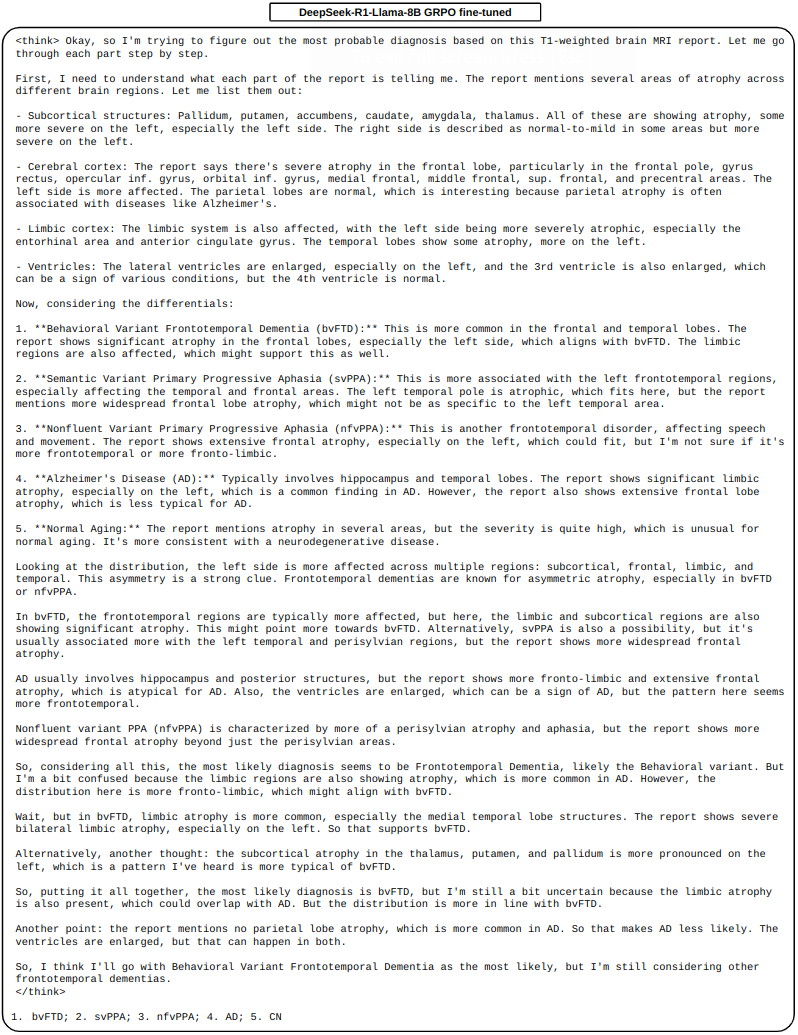}
\caption{Full diagnostic output from GRPO-trained DeepSeek-R1-Llama-8B on a bvFTD case.}
\label{fig:full_deepseek_BV}
\end{figure}

\begin{figure}[h]
\centering
\includegraphics[width=\textwidth]{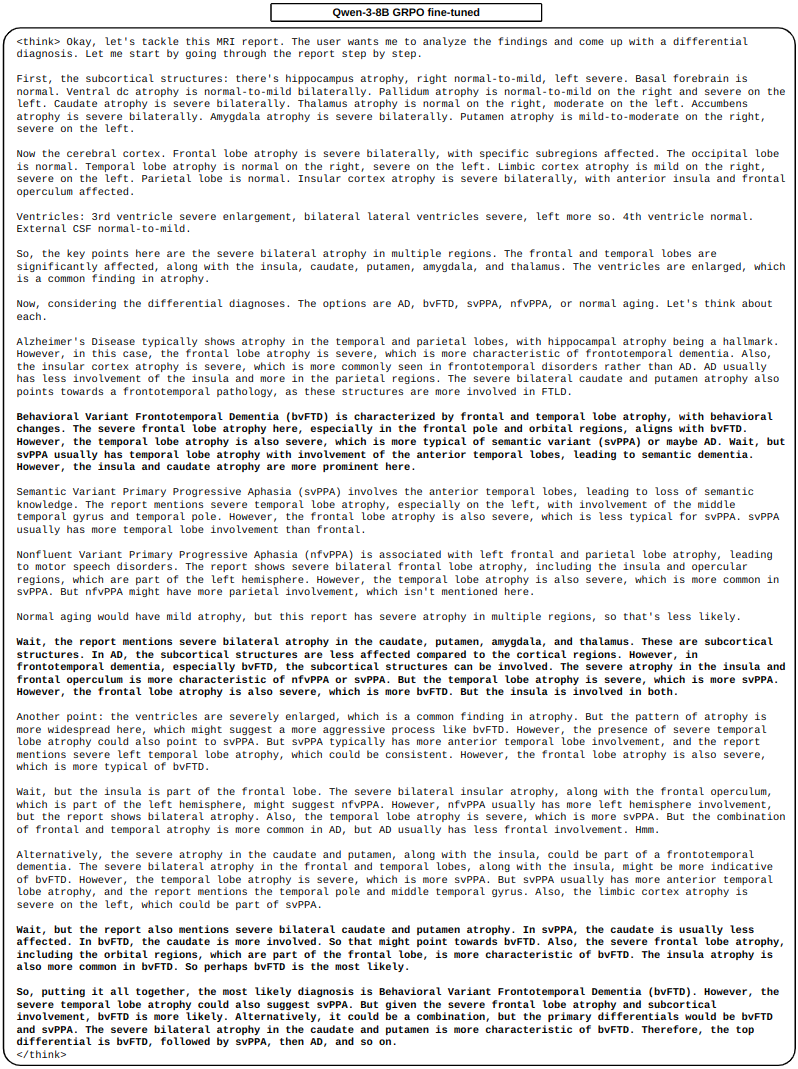}
\caption{Full diagnostic output from the GRPO-trained Qwen-3-8B model on a bvFTD case.}
\label{fig:full_qwen3_BV}
\end{figure}

\begin{figure}[h]
\centering
\includegraphics[width=\textwidth]{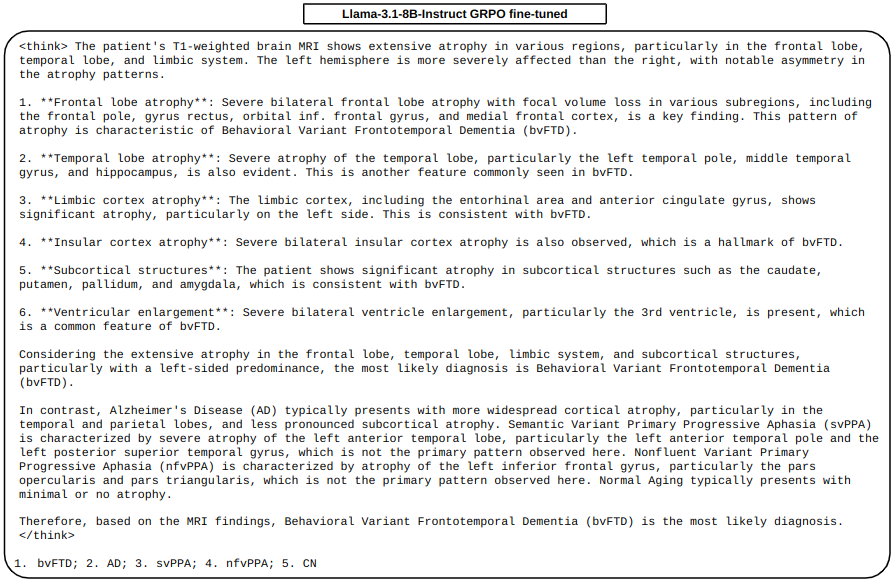}
\caption{Full diagnostic output from the GRPO-trained LlaMA-3.1-Instruct-8B model on a bvFTD case.}
\label{fig:full_llama_BV}
\end{figure}

\begin{figure}[h]
\centering
\includegraphics[width=\textwidth]{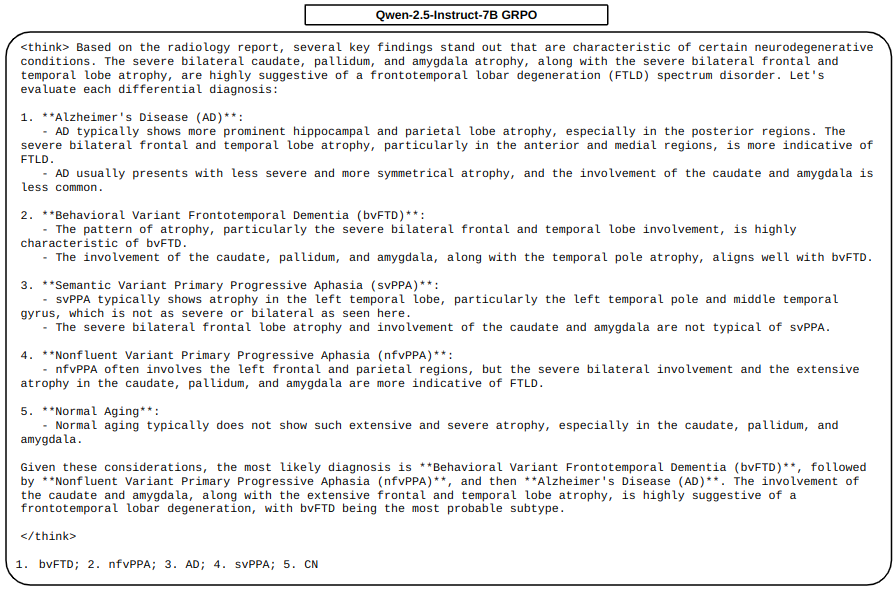}
\caption{Full diagnostic output from the GRPO-trained Qwen-2.5-Instruct-7B model on a bvFTD case.}
\label{fig:full_qwen_BV}
\end{figure}

\clearpage
\section{Training Details for Classification-Only Baselines}
\label{app:dl_training_details}

\paragraph{Vision-transformer (ViT).}
We build on the 3D ViT architecture and training pipeline from Nguyen \textit{et al.}~\cite{nguyen3DTransformerBased2023}, originally designed for AD–FTD–CN classification. To support our extended diagnostic setting, we modify the final classification head to a 5-way MLP, enabling prediction across CN, AD, bvFTD, svPPA, and nfvPPA. All other architectural components, data preprocessing steps, and optimization settings (e.g., learning rate, batch size, data augmentation) are retained from the original implementation.

\paragraph{Support Vector Machine (SVM).}
We train SVM classifiers using the Structural Deviation Score features described in Subsection~\ref{subsec:brain_to_text}. Hyperparameter tuning is performed via grid search across two kernel types (\texttt{linear}, \texttt{rbf}) and 100 logarithmically spaced values of the regularization strength \(C \in [10^{-4}, 10^{1.5}]\). The best model is selected based on balanced accuracy on the held-out validation folds.

\end{document}